\documentclass[conference]{IEEEtran}
\usepackage{caption}
\makeatletter
\def\ps@IEEEtitlepagestyle{%
  \def\@oddfoot{}%
  \def\@evenfoot{}%
}

\IEEEoverridecommandlockouts

\usepackage{array}
\usepackage{caption}
\usepackage{multirow}
\usepackage{float}
\usepackage{cite}
\usepackage{amsmath,amssymb,amsfonts}
\usepackage{algorithmic}
\usepackage{graphicx}
\usepackage{textcomp}
\usepackage{xcolor}
\usepackage{booktabs}
\usepackage{stfloats}
\usepackage{makecell}
\usepackage{rotating}
\usepackage{amsmath}

\def\BibTeX{{\rm B\kern-.05em{\sc i\kern-.025em b}\kern-.08em
    T\kern-.1667em\lower.7ex\hbox{E}\kern-.125emX}}
    
\usepackage{eso-pic}
\usepackage[colorlinks=true, urlcolor=blue, linkcolor=blue, citecolor=blue,pdfborder={0 0 0}]{hyperref}   

\begin{document}

\title{\vspace*{1cm}Universal Laboratory Model: prognosis of abnormal
clinical outcomes based on routine tests\\
}

\author{
\IEEEauthorblockN{Pavel Karpov}
\IEEEauthorblockA{\textit{Department of medicinal chemistry} \\
\textit{Moscow State University}\\
Moscow, Russia \\
carpovpv@qsar.chem.msu.ru}
\and
\IEEEauthorblockN{Ilya Petrenkov}
\IEEEauthorblockA{\textit{Department of computer science} \\
\textit{Samara National Research University}\\
Samara, Russia \\
petilia2002@yandex.ru}

\and

\IEEEauthorblockN{Ruslan Raiman}
\IEEEauthorblockA{\textit{R\&D in medical laboratories} \\
\textit{Ltd, RosLIS}\\
Moscow, Russia \\
rulsan.raiman@roslis.ru}
}

\maketitle

\begin{abstract}
Clinical laboratory results are ubiquitous in any diagnosis making. Predicting abnormal values of not prescribed tests based on the results of performed tests looks intriguing, as it would be possible to make early diagnosis available to everyone. The special place is taken by the Common Blood Count (CBC) test, as it is the most widely used clinical procedure. Combining routine biochemical panels with CBC presents a set of test-value pairs that varies from patient to patient, or, in common settings, a table with missing values. Here we formulate a tabular modeling problem as a set translation problem where the source set comprises pairs of GPT-like label column embedding and its corresponding value while the target set consists of the same type embeddings only. The proposed approach can effectively deal with missing values without implicitly estimating them and bridges the world of LLM with the tabular domain. Applying this method to clinical laboratory data, we achieve an improvement up to 8\% AUC for joint predictions of high uric acid, glucose, cholesterol, and low ferritin levels.
\end{abstract}


\section{Introduction}

The World Health Organization places the complete blood count (CBC) test in the list of basic laboratory tests \cite{WHO2021}, making it one of the most widely used diagnostic procedures. As the CBC test could be treated as a proxy \cite{SantosSilva2024,CBCCardio} to determine the pathological states of a patient, modeling these states based on the results of the CBC and other laboratory tests \cite{Kurstjens2022, Luo2016,Pullakhandam2024,Cardozo2022, Cichosz2022,Zhang2024, Lee2019, Gao2021} could early diagnose chronic diseases, preventing them from developing and improving overall well-being at the individual and state levels. 
Although much effort is being made to develop large language models to analyze textual information in medical documentation\cite{llmmedicinereview}, the use of ordinary tabular data with laboratory results is less common due to the lack of data and regular difficulties in tabular modeling\cite{babenkoTabular}. In this work, we focus on modeling abnormal values of one test (namely: ferritin\cite{Kurstjens2022, Luo2016}, uric acid\cite{Zhang2024, Lee2019, Gao2021}, cholesterol and glucose\cite{Cardozo2022})\footnote{The choice of tests is dictated by the data availability.} based solely on demographic data (gender and age) and the results of other tests (CBC and biochemical tests, if available). Thus, if a patient has CBC results, with the help of the presented approach and models, she could also have a prognosis of low ferritin levels (anemia), high uric acid (gout), high cholesterol (risk of cardiovascular disorders), and high glucose (diabetes).

Most studies explore relatively small datasets from one hospital and use a fixed set of tests (features, indicators) for each sample (classic tabular data) to develop predictive models. As a result, these datasets often have a predominant proportion of sick patients, resulting in remarkably good statistics, Table \ref{tab:overview}.

Low ferritin levels are predicted in\cite{Kurstjens2022} to detect iron deficiency anemia (IDA) using random forest-based algorithms (RF), achieving the area under the ROC curve (AUC) of 0.92. Reference \cite{Pullakhandam2024} reports the application of the gradient boost method (GB) to detect IDA and the best model reaches the AUC of 0.87, for their initial dataset, and even higher 0.89 for the so-called Kenyan dataset.

Detection of the high level of glycated hemoglobin (HbA1c) in \cite{Cardozo2022} with the help of an artificial neural network (ANN) achieves 78.1\% sensitivity, 78.7\% precision, and an F1 score of 78.4\% in diabetes diagnosis. In \cite{Cichosz2022}, the high-level HbA1c level is modeled, resulting in an AUC of 0.81.

Models for predicting high uric acid values are presented in \cite{Zhang2024}, where the stacking ensemble model (SEM) reveals the best results: 0.91 accuracy, 0.89 sensitivity, 0.91 specificity, and an F1 score of 0.65. The authors \cite{Sampa2020} propose a regression model for uric acid, achieving a determination coefficient of 0.99.

\setlength{\abovecaptionskip}{2pt}
\setlength{\belowcaptionskip}{1pt}

\renewcommand{\arraystretch}{1.5}
\begin{table*}[t]
\caption{An overview of laboratory data modeling}
\label{tab:overview}
\begin{center}
\setlength{\tabcolsep}{4pt}
\begin{tabular}{m{2.7cm}rm{3.5cm}cm{7.8cm}c}
\toprule
\textbf{Test} & \textbf{Data size} & \textbf{Thresholds$^{\mathrm{a}}$} & \textbf{AUC$^{\mathrm{b}}$} & \textbf{Methods} & \textbf{Refs} \\
\midrule
Glycated\newline hemoglobin, HbA1c & 45\,431 &  $\ge$6,5\% & 0.81 & AdaBoost, ANN, LogitBoost, RF, RUSBoost & \cite{Cichosz2022} \\
Hemoglobin, (HGB) & 364 & $\le$130 g/L for men (M),\newline $\le$120 g/L for women (F) & 0.97 &ANN, Decision Tree (DT), Logistic Regression (LR), NaïveBayes (NB), RF, Support Vector Machine (SVM) & \cite{Vohra2022} \\
Ferritin, (FER) & 12\,009 & $\le10-13 \,\mu\text{g/L}$ for women,\newline $\le22-30 \,\mu\text{g/L}$ for men$^{\mathrm{c}}$ & 0.92 & RF & 
\cite{Kurstjens2022} \\
Ferritin, (FER) & 5\,128 & $\le10 \,\mu\text{g/L}$ for women\newline $\le30 \,\mu\text{g/L}$ for men & 0.97 & Bayesian linear regression, Lasso regression, Linear regression, RF & \cite{Luo2016} \\
Irion deficiancy\newline anemia, (IDA) & 19\,000 & $\le15 \,\mu\text{g/L}$ & 0.87 &  GB, K-Nearest Neighbors (KNN), LR, NB, RF, XGBoost (XGB) & \cite{Pullakhandam2024} \\
Uric Acid, (URIC) & 40\,899 & $\ge0,42$ mmol/L for men\newline $\ge0,36$ mmol/L for women & 0.85 & DT, SEM, SVM, XGB & \cite{Zhang2024} \\
Uric Acid, (URIC) & 91\,690 & ---\texttt{"}---  & 0.82 & LR, RF & \cite{Gao2021} \\
Uric Acid, (URIC) & 38\,001 &  ---\texttt{"}--- & 0.78 & Discrimination analysis classification, DT, KNN, NB, RF, SVM  & \cite{Lee2019} \\
\bottomrule
\multicolumn{6}{p{17.7cm}}{$^{\mathrm{a}}$If the condition is true, the point belongs to the positive class. $^{\mathrm{b}}$Best AUC values are reported if there are many methods used in the article. $^{\mathrm{c}}$ The authors used different thresholds depending on the laboratory equipment.}
\end{tabular}
\end{center}
\vspace{-0.5cm}
\end{table*}

The use of fixed sets of tests to build a model contradicts common clinical practice, where a set of prescribed tests for one patient would likely be different from the set of tests for another patient as the choice is usually dependent on the doctors, patients' jobs, individual health state, etc. To improve the accuracy of the predictions, it is important to consider all available information. Therefore, the input data for the predictive models in this case is a sparse matrix, Table \ref{tab:data}. This leads to the problem of "missing values", where different patients lack different tests, complicating the application of standard machine learning methods.

There are many approaches to handling missing values, including machine learning-based methods \cite{emmanuel2021survey}, mean imputation \cite{joel2022review}, genetic algorithms \cite{katoch2021review, sulejmani2021development}, Generative Adversarial Networks (GANs) \cite{yoon2020gamin, kim2020survey}, etc.

For example, in\cite{Luo2016}, the authors restore the missing values before predicting low ferritin levels. The four imputation techniques were used: mean, multiple imputation with chained equations–full (MICE-full), multiple imputation with chained equations–select (MICE-sel) and missForest. In the articles \cite{Vohra2022, el2019machine}, the authors predict the missing hemoglobin values (make a regression) before determining the type of anemia.

All imputation methods rely on assumptions about the data distribution, inevitably leading to distortions. Laboratory results are individual for each patient. This uniqueness should be carefully considered when developing machine learning models. A more effective approach may be to adapt the predictive model to handle missing values rather than trying to fill them in.

Our contributions in the article are as follows:

\begin{itemize}

\item\rule{0.cm}{0.5cm}We developed a new neural network architecture -- the Universal Laboratory Model (ULM) -- that can effectively model multidimensional laboratory data with missing values using the attention mechanism of the Set Transformer \cite{SetTransformer} and internal encodings of key parameters by the Large Language Model \cite{Vaswany, yandexgpt}.

\item\rule{0.cm}{0.5cm}Applying this architecture to laboratory data, we built accurate models to predict system pathology that can result in low ferritin, high uric acid, cholesterol, and glucose levels based on CBC results and (if any) other blood tests. The model can be easily expanded for other tests (features).

\item\rule{0.cm}{0.5cm}We made these models publicly available on \href{https://ulm.roslis.ru}{https://ulm.roslis.ru} to facilitate their use by Laboratory Information Systems throughout the world. The source code developed in the work is also available on GitHub, \href{https://github.com/petilia2002/universal-laboratory-model}{https://github.com/petilia2002/universal-laboratory-model}.

\end{itemize}

\section{Methods}
\subsection{Dataset}

Data used for this study were collected from drivers connecting the Laboratory Information System (RosLIS\footnote{\href{https://roslis.ru}{https://roslis.ru}}) to analytical equipment in a number of commercial medical laboratories in Russia. These laboratories provide service for ordinary people, patients in hospitals, check-up inspections, making the data quite diverse and not biased to a population of narrow specialized organizations as is used in many contemporary studies; see Table \ref{tab:overview}. The size of our database is 1\,173\,403 data rows and is 10 times larger compared to known studies.

As the data came from laboratories that analyze several thousands samples per day, technical errors (test samples, equipment, or quality control failures, not human samples (animals, dead organisms, etc.) are inevitable and a special procedure to eliminate clearly wrong values is needed as there are no standard intervals to exclude outliers taking into account that a severe pathology may have been observed. By developing this procedure, we also keep in mind an applicability domain estimation of the model to ensure that it can provide reliable results.

\subsection{Applicability domain}

Any mathematical model can only produce reliable results if the input data fall in the model's applicability domain (AD), which usually includes ranges for all continuous parameters and predicted outcomes. Acceptable value ranges can be determined for each blood parameter by constructing a probability density function that captures variations in the parameter. The acceptable range is then set to include some predefined percentage of the area under the distribution curve. As our database has more than one million rows and the data are supposed to be reliable, we set this threshold to a high value, 99.99\%. 

Our implementation of this algorithm includes building a histogram of all the values of a test, sorting it by them, finding the maximum, and traversing from this point in both directions until the fraction of the cumulative number of visited values is less than the threshold.

The ranges found are presented in Table \ref{tab:areas}. Interestingly, the minimum acceptable value for HGB is 11 g/L as it is known that less than 10 g/L for the HGB test is not compatible with life. In contrast, the maximum value of NEUT is 100\% seems to be unreliable since NEUT is a part of WBC and very rare a part could be the whole. We decided not to exclude these values because, from a numerical perspective, they would not induce any distortion. 

In addition to ranges, there is also another condition that must be satisfied for a sample to be in the AD - the sum of all WBC parts from a 5-DIFF analyzer, namely BASO, EOS, LYMPH, MONO, and NEUT (or MID, GRA, and LYMP for 3-DIFF) must equal \(100 \pm 4 \%\)\footnote{We can accept a small difference from 100\% as sometimes the values are edited by hand.}.

\setlength{\abovecaptionskip}{5pt}
\setlength{\belowcaptionskip}{0pt}
\renewcommand{\arraystretch}{1.5}
\setlength{\tabcolsep}{3pt}

\begin{table}[t]
\centering
\caption{Ranges of values of parameters, alphabetical order}
\label{tab:areas}
\begin{tabular}{c>{\raggedright}p{3.4cm}rrp{0.9cm}p{1.3cm}}
    \toprule
    \textbf{\#} & \textbf{Feature (Test)} & \textbf{Min} & \textbf{Max} & \textbf{Units} & \textbf{LOINC$^{\mathrm{a}}$} \\
    \midrule
    1  & Age & 1 & 110 & years & -- \\    
    2 & Alanine transaminase, (ALT) & 0.10 & 1721.40 & U/L & \href{https://loinc.org/1742-6}{1742-6} \\
    3 & Albumin, (ALB) & 0.04 & 60.14 & g/L & \href{https://loinc.org/1751-7}{1751-7} \\
    4 & Aspartate aminotransferase, (AST) & 0.30 & 1518.50 & U/L & \href{https://loinc.org/1920-8}{1920-8} \\
    5 & Basophils, (BASO) & 0.02 & 27.20 & \% & \href{https://loinc.org/704-7}{704-7} \\
    6 & Cholesterol, (CHOL) & 0.02 & 17.20 & mmol/L & \href{https://loinc.org/14647-2}{14647-2} \\
    7 & Creatinine, (CREA) & 0.30 & 1618.80 & $\mu$mol/L & \href{https://loinc.org/14682-9}{14682-9} \\
    8 & C-reactive protein, (CRP) & 0.01 & 250.95 & mg/L & \href{https://loinc.org/1988-5}{1988-5} \\
    9 & Direct bilirubin, (BC) & 0.04 & 238.40 & $\mu$mol/L & \href{https://loinc.org/29760-6}{29760-6} \\
    10 & Eosinophils, (EOS) & 0.09 & 43.80 & \% & \href{https://loinc.org/711-2}{711-2} \\
    11 & Ferritin, (FER) & 0.01 & 1667.20 & $\mu\mbox{g/L}$ & \href{https://loinc.org/20567-4}{20567-4}\\
    12 & Folic acid, (FOL) & 0.56 & 330.11 & ng/mL & \href{https://loinc.org/2284-8}{2284-8} \\   
    13 & Gender, ($\{M=1, F= 0\}$) & 0 & 1 & -- & -- \\ 
    14 & Glucose, (GLU) & 0.01 & 26.77 & mmol/L & \href{https://loinc.org/14771-0}{14771-0} \\
    15 & Granulocytes, (GRA) & 14.7 & 94.7 & \% & \href{https://loinc.org/19023-1}{19023-1}  \\
    16 & Hemoglobin, (HGB) & 11.00 & 215.00 & g/L & \href{https://loinc.org/30350-3}{30350-3} \\
    17 & Indirect bilirubin, (BU) & 0.05 & 220.79 & $\mu$mol/L & \href{https://loinc.org/14630-8}{14630-8} \\
    18 & Lactate dehydrogenase, (LDH) & 2.00 & 4983.00 & U/L & \href{https://loinc.org/2532-0}{2532-0} \\
    19 & Lymphocytes, (LYMPH) & 0.10 & 90.70 & \% & \href{https://loinc.org/737-7}{737-7} \\
    20 & Mean corpuscular volume, (MCV) & 0.70 & 134.00 & fL & \href{https://loinc.org/71829-6}{71829-6} \\
    21 & Middle-size Cells, (MID) & 1.2 & 29.4 & \% & \href{https://loinc.org/32155-4}{32155-4}\\ 
    22 & Monocytes, (MONO) & 0.10 & 55.40 & \% & \href{https://loinc.org/5905-5}{5905-5} \\
    23 & Neutrophils, (NEUT) & 2.90 & 100.00 & \% & \href{https://loinc.org/768-2}{768-2} \\
    24 & Platelets, (PLT) & 1.00 & 1053.00 & $10^{9}$/L & \href{https://loinc.org/777-3}{777-3} \\
    25 & Red blood cells, (RBC) & 0.24 & 8.21 & $10^{12}$/L & \href{https://loinc.org/704-7}{789-8} \\
    26 & Total bilirubin, (TBIL) & 0.03 & 432.43 & $\mu$mol/L & \href{https://loinc.org/704-7}{54363-7} \\
    27 & Total protein, (PRO) & 19.20 & 132.10 & g/L & \href{https://loinc.org/704-7}{13980-8} \\
    28 & Urea, (UREA) & 0.50 & 67.50 & mmol/L & \href{https://loinc.org/22664-7}{22664-7} \\
    29 & Uric acid, (URIC) & 0.00 & 1.22 & mmol/L & \href{https://loinc.org/14933-6}{14933-6} \\
    30 & Vitamin B12, (VB) & 1.00 & 39833.00 & pg/mL & \href{https://loinc.org/2132-9}{2132-9} \\
    31 & White blood cells, (WBC) & 0.10 & 68.70 & $10^{9}$/L & \href{https://loinc.org/804-5}{804-5} \\
    \bottomrule
    \multicolumn{6}{p{8.6cm}}{$^{\mathrm{a}}$Logical Observation Identifiers Names and Codes -- \href{https://loinc.org/get-started/what-loinc-is/}{https://loinc.org}} 
\end{tabular}
\vspace{-0.5cm}
\end{table}

\subsection{Universal Laboratory Model}

Building a model on tabular data lacks information on the data itself. All the model knows is that there are some columns and that there are some values in these columns. But it has no clue what exactly each column means and how these columns are related or what objectives of the model are and how they possibly can relate between each other. In general, there is no context that the model can benefit from when training for a particular task. Without this context, deep learning models stagnate and are usually not better than classical machine learning algorithms like RF or SVM. 

\begin{figure}[t]
    \centering
    \includegraphics[width=1\linewidth]{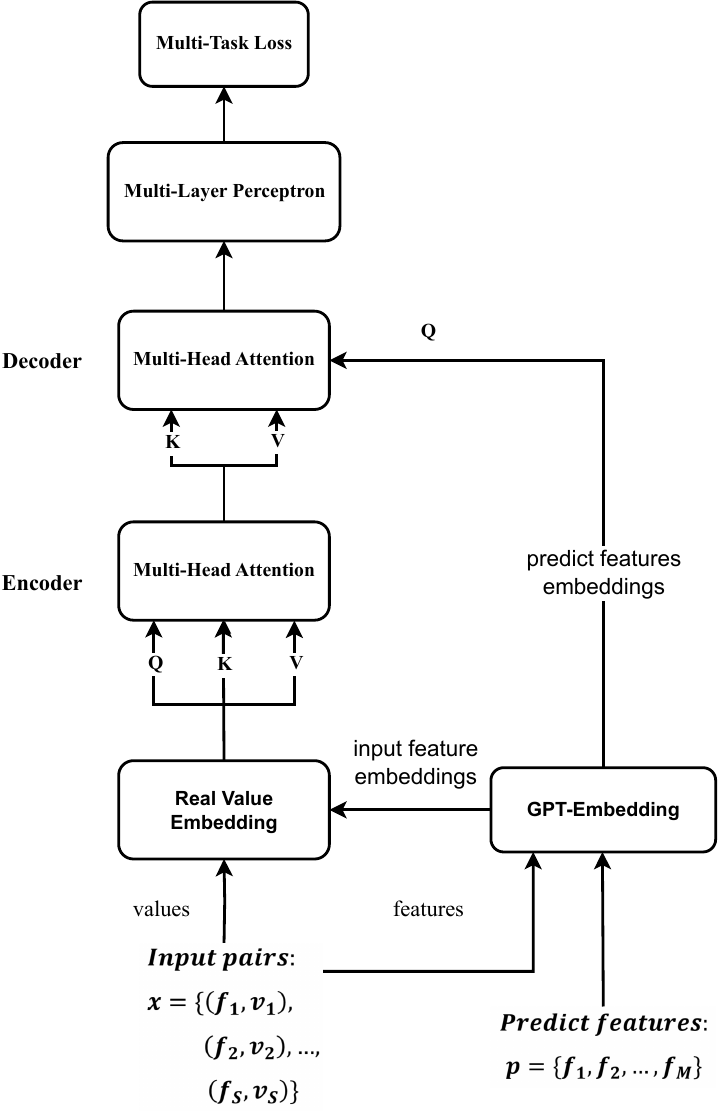}
    \caption{ULM architecture: Encoder-Decoder pattern, Decoder uses predicting feature embeddings as queries (Q)}
    \label{fig:ulm}
\end{figure}

To catch up on these relations between input and output and provide context, an attention mechanism can be used. Similar column embeddings were introduced in \cite{TabTransformer} as trainable parameters that can improve the performance of the model. Instead, we use GPT-like embeddings of features that are fixed and propose to use the same-type embeddings for objectives of the model as well. 
Our approach uses the Encoder-Decoder pattern, Fig. \ref{fig:ulm}. The attention block has three parameters, namely queries $Q$, keys $K$, and values $V$\cite{Vaswany}. If $X$ denotes input data and $P$ -- predicting features, then for the encoder $E(X)$, we use the so-called Self-Attention, where all the parameters are the same and correspond to the input of the model; for the decoder $D(P,X)$ the $Q=P$ and comes from the features the model is training to predict:
\begin{align}
E(X) = Attention(Q=X, K=X, V=X) \\
D(P, X) = Attention(Q=P, K=I, V=I)
\end{align}

There is no natural order in placing test results on a sheet of a laboratory report except traditions and perhaps state regulations. If one casts away missing values, all position information is gone, and all is rest is a number of unordered sets. To convert a set to a vector suitable for machine learning, one has to make sure that an order of elements in this vector does not affect model's output. 

Pooling techniques such as dimension-wise average or maximum of feature vectors are the usual choice to make this output invariant. The attention block (multi-head attention as well)\cite{Vaswany} is equivariant, meaning that any permutation in the input gives the same permutation in the output. To make it invariant, it is possible to utilize a multi-head attention pooling\cite{SetTransformer} induced by queries $Q$, which are not dependent on the order of input and, therefore, could be learnable. By construction, ULM can work with sets of different sizes and orders and produce a consistent prognosis.

For simplicity, we consider only one laboratory report (for one patient). Let $x = (f, v)$ denote a pair of feature values (test result), where $f$ is a categorical feature of a set of all available features (tests), Table \ref{tab:areas} (column \#), $F = \{f_1, f_2, \ldots, f_L\}$, including age, gender, and laboratory tests names\footnote{Strictly speaking, age and gender have a different nature compared to laboratory tests though they could be examined and, therefore, treated as results as well.}, $L$ - number of all possible features (here 31); $v \in \mathbb{R}$ - is a numerical value of that feature. A subset $P \subset F$ contains the tests which abnormal values the model is going to predict. As we restrict ourselves to model only for tests (GLU, CHOL, FER, URIC), the cardinality of $P$ is 4. A laboratory report consists of $O$ results and could be expressed as $\mathbf{r} = \{x_1, x_2, \ldots, x_O\}$, where $O\le L$. Among these $O$ pairs, we treat $S$ pairs as training input data $\mathbf{x} = \{x_1, x_2, \ldots, x_S \mid f \not\in P\}$ and $S\le L - |P|$ and the rest of $\mathbf{r}$ is training output data $\mathbf{y} =\{\phi_1(x_1), \phi_2(x_2), \ldots, \phi_M(x_M) \mid f \in P\}$, where $M=|P|$ and $\phi_i$ is a function that converts a real-value of a test $i$ into a binary class as in Table \ref{tab:results}. 

Each row in Table \ref{tab:data} corresponds to $\mathbf{r}$; $\mathbf{x}$ is a part of $\mathbf{r}$ placed to the left from the vertical line, and the source for $\mathbf{y}$ is a part of $\textbf{r}$ placed to the right from this line. For sample \#1 in Table \ref{tab:data}:
\begin{equation*}
\begin{aligned}
\mathbf{x}_1 ={} & \{ (13, 0), (1, 55), (16, 128), (25, 4.48) \\
                 &    (20, 85), (28, 3.4), (7, 75.2) \}  \\
\mathbf{y}_1 ={} & \{ (14, 0), (11, 1), (29, 0) \}             
\end{aligned}
\end{equation*}

For different samples (patients), the $O$ may vary as the doctors order different sets of tests depending on the patient's state, possible disease, or general objective of an inspection. Moreover, the training set is constructed so $\mathbf{y} \not\in \varnothing$, while during inference $\mathbf{y}$ must be empty or at least not equal to $P$.

\begin{table}[t]
\caption{Example records from our database, $N$ = 1\,173\,403. Data to the left of the line is used as input, to the right -- output. Empty cells depict missing values.}
    \centering
    \setlength\rotheadsize{0.92cm}
    \setlength{\tabcolsep}{0.40em}
    \begin{tabular}{ccccccccc|ccccc} \toprule
    \# & \rothead{\textbf{Gender}} & 
         \rothead{\textbf{Age}} & 
         \rothead{\textbf{HGB}} & 
         \rothead{\textbf{RBC}} & 
         \rothead{\textbf{MCV}} & 
         ... &  
         \rothead{\textbf{UREA}} & 
         \rothead{\textbf{CREA}} &          
         \rothead{\textbf{GLU}} & 
         \rothead{\textbf{CHOL}} & 
         \rothead{\textbf{FER}} & 
         \rothead{\textbf{URIC}} 
    \\ \midrule
    1  & F & 55 & 128 & 4.48 & 85 & . & 3.4 & 75.2 & 5.37 &  & 8.67 & 0.247 \\
    2  & F & 33 & 137 & 4.22 & 97.2 & . & & & 4.4 & 5.44 & 124.1 & 0.521 \\    
    3  & M & 66 & 146 & 4.29 & 100 & . & 109 &  & 8.24 & 5.29 & 94.6 & 0.314 \\
    . & . & . & . & . & . & . & . & . & .  & .  & . & . \\ 

    $N$  & M & 31 & 141 & 4.65 & 86 & . & 3.2 & 86.7 & & 3.98 &  & 0.327 \\
    \bottomrule 
    \end{tabular}    
    \label{tab:data}
    \vspace{-0.3cm}
\end{table}

We believe GPT-embeddings carry important information for the model and all rotations and squeezing of these embeddings destruct GPT knowledge, therefore, any matrix multiplication must be refrained. Only scalar multiplication and translation are left, so the output of Real Value Embeddings layer will be:
\begin{equation}
    y_i = GPT_i * v_i + B
\end{equation}
where $GPT$ stands for GPT-embeddings of feature $i$, $v_i$ is the value of that feature, $B$ is a trainable translation parameter, and $*$ is scalar multiplication. 

\subsection{Data preparation and scaling}

To form the dataset for training and testing, we apply the ranges in Table \ref{tab:areas} to all available data and select only those results that fall in the ranges and contain values of at least one of the tests from $P$, that is glucose, cholesterol, ferritin, or uric acid. For every value $v$ in $\textbf{x}$ we apply the logarithmic transformation since there are no zero values and then scale them to the interval $[0.1, 0.9]$ for normalization.

\subsection{Training and validation protocols}

All models were trained with the same protocol. The dataset was randomly divided 80\% for training and 20\% for testing. Training was controlled by the Adam optimization algorithm with early stopping and automatic reduction of the learning rate techniques. The total number of epochs was limited to 1\,000, and the statistics in Table \ref{tab:results} are the result of three different training runs for each variant. For standard Multilayer Perceptron (MLP) models, the optimization started with the learning rate $10^{-3}$, for ULM $10^{-4}$. The batch size was 32; to support batch mode, we used the usual Transformer Masks for the encoder and decoder. Tests embeddings were calculated using the open Yandex GPT service\cite{yandexgpt}. All  code is written in Python3 programming language. The Keras framework\cite{keras} for machine learning with TensorFlow\cite{tensorflow} backend is used.

\section{Results}

\begin{table}[b]
\centering
\caption{Models' AUC values (\%) (the bigger the better)}
\label{tab:results}
\begin{tabular}{m{1cm}m{2cm}>{\raggedleft}m{0.95cm}>{\centering\arraybackslash}m{1.15cm}>{\centering\arraybackslash}m{1.15cm}>{\centering\arraybackslash}m{1.15cm}} \toprule 
\textbf{Test} & \textbf{Threshold}$^{\mathrm{a}}$ & \centering\textbf{Size} & \textbf{MLP\_B}$^{\mathrm{b}}$ & \textbf{MLP\_M}$^{\mathrm{c}}$ & \textbf{ULM} \\ \midrule 
GLU & $\ge$7 mmol/L & 680\,502 & 80.2$\pm$0.1 & 80.4$\pm$0.1 & \textbf{82.4}$\pm$0.1 \\
CHOL & $\ge$5.2 mmol/L & 248\,806 & 75.7$\pm$0.1 & 75.6$\pm$0.1 & \textbf{77.9}$\pm$0.1 \\ 
FER & $\le$12 ng/mL & 105\,928 & 88.9$\pm$0.2 & 88.7$\pm$0.1 & \textbf{89.7}$\pm$0.1 \\
URIC & M:$\ge$0.48\,mmol/L\newline F:$\ge$0.38\,mmol/L& 92\,901 & 71.5$\pm$0.2 & 71.5$\pm$0.2 & \textbf{79.7}$\pm$0.2 \\ \bottomrule
\multicolumn{6}{p{8.6cm}}{$^{\mathrm{a}}$If the condition is true, the sample is treated as positive. $^{\mathrm{b}}$Binary classification: the model has one output and each analyte is modeled independently of the others. $^{\mathrm{c}}$Multiclass classification: the model has 4 outputs, e.g. all analytes are modeled simultaneously.}
\end{tabular}
\end{table}

In our first experiments, we used one output (refer to MLP\_B in Table \ref{tab:results}) to build classification models that are simple 3-layer neural networks with architecture 27 x 256 (relu) x 1 (sigmoid) and binary cross-entropy loss. We did not observe any dependency on the number of hidden neurons or activation function in the middle layer. Then we switched to multitask \cite{MultiTask} settings (MLP\_M) when the output comprises several masked values with masked categorical cross-entropy loss and the architecture is 27 x 256 (relu) x 4 (sigmoid); but no notable improvements were found. For all MLP models, we varied training settings, activation functions, normalization schemes, applied different residual connections, built numerical embeddings \cite{BabenkoEmbeddings}, performed internal ensemble \cite{ensemble}, but all our efforts did not affect the performance of the models; small fluctuations in AUC - no better, no worse. 

\begin{figure}[b]
    \centering
    \includegraphics[width=1.0\linewidth]{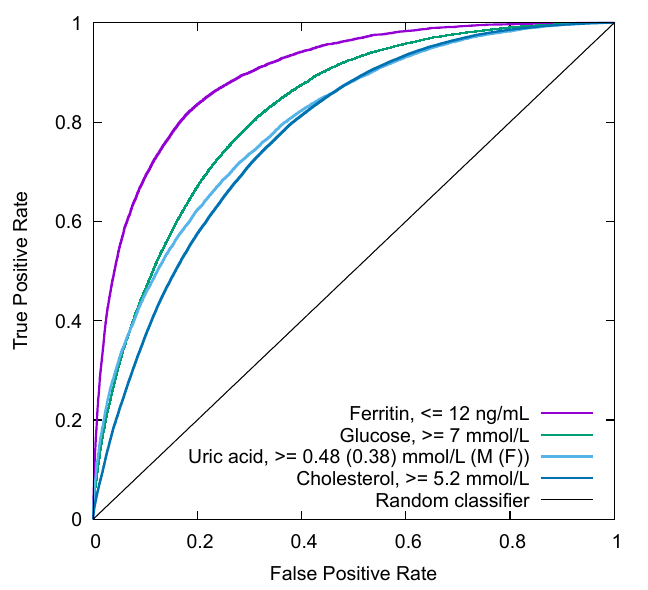}
    \caption{ROC-curves for ULM model. Positive class means prediction of abnormal value.}
    \label{fig:roc}
    \vspace{-0.3cm}
\end{figure}

The next step was to use ULM for the same task. ULM uses a multi-head attention encoder and decoder with a number of heads equal to 8 and key dimension 16. The dropout is set to 0.1. ULM has the same output and loss as MLP\_M. ULM improved the statistics for all modeled tests, Table \ref{tab:results}. The ROC curves for the ULM model are shown in Fig. \ref{fig:roc}, confusion matrix pesented in Table \ref{tab:spec}. 

\begin{table}[h]
    \centering
    \caption{Parameters of the ULM model}
    \begin{tabular}{p{2.8cm}>{\raggedleft}p{1.1cm}>{\raggedleft}p{1.3cm}>{\raggedleft}p{1.1cm}>{\raggedleft\arraybackslash}p{1.4cm}} \toprule 
       \textbf{Parameter}  & \textbf{Glucose} & \textbf{Cholesterol} & \textbf{Ferritin} &  \textbf{Uric acid}\\ \midrule 
       True Positive (TP) & 5\,337 & 22\,616 & 2\,066 & 1\,940 \\ 
       True Negative (TN) & 101\,194 & 16\,276 & 16\,488 & 12\,100 \\ 
       False Positive (FP) & 34\,454 & 6\,928 & 3\,850 & 4\,965 \\
       False Negative (FN) & 1\,943 & 9\,657 & 520 & 901 \\ \midrule
       Accuracy (\%) & 74.5 & 70.1 & 80.9 & 70.5\\
       Sensitivity (\%) & 73.3 & 70.1 & 79.9 & 68.3\\ 
       Specificity (\%) & 74.6 & 70.1 & 81.1 & 70.9 \\ 
       \bottomrule 
    \end{tabular}    
    \label{tab:spec}
    \vspace{-0.3cm}
\end{table}

The improvement for Ferritin is the smallest 0.8\% (but significant), as this test has a strong relation to only CBC (especially HGB) and, as we see, additional biochemical data does not substantially change the situation. The AUC for Glucose and Cholesterol outputs got $\sim$ 2\% gain, while the increase for Uric acid jumps up to 8\%. One possible explanation is that high values of Uric acid in the blood cause gout, and during attack a patient has severe pain in the joints that affects all functions of the organisms and is reflected in the change in CBC and other biochemical tests.

\section{Conclusions and outlook}

In this article, we propose the Universal Laboratory Model that uses context GPT-like embeddings for both input and output and can effectively work with different size sets thanks to the invariant possibility in the encoder-decoder architecture. In four laboratory cases, we demonstrate that the use of ULM can improve model performance compared to standard MLP, and the models can be used in laboratory diagnostics.

Still there are open questions. What is a right threshold to mark a positive case? As shown in Table \ref{tab:overview}, different groups define different thresholds leading to inconsistent statistics. Moreover, every reading has its own error and is influenced by more broad biological variation. We think incorporation of this additional data could also be beneficial for the models making them more consistent. 

Another problem is data. As we stress, many groups use a limited number of data points to build and compare models. Usually the origin of these data is one hospital and therefore is subjected to limited groups of patient that already have some kind of health disorders, so the data contains more positive examples than negative and the model statistics are biased to positive cases showing remarkably good values. Utilizing datasets from different hospitals and even countries raise more complex tasks: How to deal with state regulations that personal data (including any medical records) originating within a country must stay within this country? Techniques such as federated learning \cite{federated} might help, but despite the success in theory and even practice in decentralized computation and data storage, its use is challenging and limited. 

From the patient's perspective, the question "what to do with a prognosis in practice" arises. Simple commenting on the result on a sheet of laboratory report like "You may have high uric acid" is useless, as it brings more questions and concerns to the patient than helps her, especially if the prognosis is wrong. Most physicians would not recommend such comments. Another possible solution that we propose is to inform the physician through any channels of integration between medical information systems (MIS). In this scenario, when a doctor views a laboratory report, MIS sends data to the AI service and presents its recommendation, but the final action depends on the doctor. 

Finally, we suggest that laboratories use these models to automatically add new tests that are predicted as positive outcomes. This would work as follows; when a laboratory system receives results, it asks for the AI service, and if there are positive predictions and biological material from the same patient is also available, the corresponding tests are added and performed automatically. In this scenario, the patient receives a laboratory report with additional test results that were not prescribed, and the doctor can start appropriate treatment the very next time she sees the patient.

\bibliographystyle{IEEEtran}
\bibliography{IEEEabrv, bibliography}   

\begin{thebibliography}{10}
\providecommand{\url}[1]{#1}
\csname url@samestyle\endcsname
\providecommand{\newblock}{\relax}
\providecommand{\bibinfo}[2]{#2}
\providecommand{\BIBentrySTDinterwordspacing}{\spaceskip=0pt\relax}
\providecommand{\BIBentryALTinterwordstretchfactor}{4}
\providecommand{\BIBentryALTinterwordspacing}{\spaceskip=\fontdimen2\font plus
\BIBentryALTinterwordstretchfactor\fontdimen3\font minus
  \fontdimen4\font\relax}
\providecommand{\BIBforeignlanguage}[2]{{%
\expandafter\ifx\csname l@#1\endcsname\relax
\typeout{** WARNING: IEEEtran.bst: No hyphenation pattern has been}%
\typeout{** loaded for the language `#1'. Using the pattern for}%
\typeout{** the default language instead.}%
\else
\language=\csname l@#1\endcsname
\fi
#2}}
\providecommand{\BIBdecl}{\relax}
\BIBdecl

\bibitem{WHO2021}
\BIBentryALTinterwordspacing
{World Health Organization}, ``The selection and use of essential in vitro
  diagnostics: Report of the third meeting of the strategic advisory group of
  experts on in vitro diagnostics,'' World Health Organization, Tech. Rep.
  1031, 2021. [Online]. Available:
  \url{https://iris.who.int/bitstream/handle/10665/339064/9789240019102-eng.pdf?sequence=1}
\BIBentrySTDinterwordspacing

\bibitem{SantosSilva2024}
\BIBentryALTinterwordspacing
M.~A. Santos-Silva, N.~Sousa, and J.~C. Sousa, ``Artificial intelligence in
  routine blood tests,'' \emph{Frontiers in Medical Engineering}, vol.~2, 2024.
  [Online]. Available:
  \url{https://www.frontiersin.org/journals/medical-engineering/articles/10.3389/fmede.2024.1369265}
\BIBentrySTDinterwordspacing

\bibitem{CBCCardio}
\BIBentryALTinterwordspacing
I.-H. Seo and Y.~Lee, ``Usefulness of complete blood count (cbc) to assess
  cardiovascular and metabolic diseases in clinical settings: A comprehensive
  literature review,'' \emph{Biomedicines}, vol.~10, 2022. [Online]. Available:
  \url{https://api.semanticscholar.org/CorpusID:253148030}
\BIBentrySTDinterwordspacing

\bibitem{Kurstjens2022}
\BIBentryALTinterwordspacing
S.~Kurstjens, T.~de~Bel, A.~van~der Horst, R.~Kusters, J.~Krabbe, and J.~van
  Balveren, ``\BIBforeignlanguage{english}{Automated prediction of low ferritin
  concentrations using a machine learning algorithm},''
  \emph{\BIBforeignlanguage{english}{Clinical Chemistry and Laboratory Medicine
  (CCLM)}}, vol.~60, no.~12, pp. 1921--1928, 2022. [Online]. Available:
  \url{https://doi.org/10.1515/cclm-2021-1194}
\BIBentrySTDinterwordspacing

\bibitem{Luo2016}
\BIBentryALTinterwordspacing
Y.~Luo, P.~Szolovits, A.~S. Dighe, and J.~M. Baron, ``Using machine learning to
  predict laboratory test results,'' \emph{American Journal of Clinical
  Pathology}, vol. 145, no.~6, pp. 778--788, 06 2016. [Online]. Available:
  \url{https://doi.org/10.1093/ajcp/aqw064}
\BIBentrySTDinterwordspacing

\bibitem{Pullakhandam2024}
\BIBentryALTinterwordspacing
S.~Pullakhandam and S.~McRoy, ``{C}lassification and {E}xplanation of {I}ron
  {D}eficiency {A}nemia from {C}omplete {B}lood {C}ount {D}ata {U}sing
  {M}achine {L}earning,'' \emph{BioMedInformatics}, vol.~4, no.~1, pp.
  661--672, 2024. [Online]. Available:
  \url{https://www.mdpi.com/2673-7426/4/1/36}
\BIBentrySTDinterwordspacing

\bibitem{Cardozo2022}
\BIBentryALTinterwordspacing
G.~Cardozo, G.~B. Pintarelli, G.~R. Andreis, A.~C.~W. Lopes, and J.~L.~B.
  Marques, ``Use of machine learning and routine laboratory tests for diabetes
  mellitus screening,'' \emph{BioMed Research International}, vol. 2022, no.~1,
  p. 8114049, 2022. [Online]. Available:
  \url{https://onlinelibrary.wiley.com/doi/abs/10.1155/2022/8114049}
\BIBentrySTDinterwordspacing

\bibitem{Cichosz2022}
\BIBentryALTinterwordspacing
S.~L. Cichosz, C.~Bender, and O.~Hejlesen, ``A comparative analysis of machine
  learning models for the detection of undiagnosed diabetes patients,''
  \emph{Diabetology}, vol.~5, no.~1, pp. 1--11, 2024. [Online]. Available:
  \url{https://www.mdpi.com/2673-4540/5/1/1}
\BIBentrySTDinterwordspacing

\bibitem{Zhang2024}
\BIBentryALTinterwordspacing
Y.~Zhang, L.~Zhang, H.~Lv, and G.~Zhang, ``Ensemble machine learning prediction
  of hyperuricemia based on a prospective health checkup population,''
  \emph{Frontiers in Physiology}, vol.~15, 2024. [Online]. Available:
  \url{https://www.frontiersin.org/journals/physiology/articles/10.3389/fphys.2024.1357404}
\BIBentrySTDinterwordspacing

\bibitem{Lee2019}
\BIBentryALTinterwordspacing
S.~Lee, E.~K. Choe, and B.~Park, ``Exploration of machine learning for
  hyperuricemia prediction models based on basic health checkup tests,''
  \emph{Journal of Clinical Medicine}, vol.~8, no.~2, 2019. [Online].
  Available: \url{https://www.mdpi.com/2077-0383/8/2/172}
\BIBentrySTDinterwordspacing

\bibitem{Gao2021}
\BIBentryALTinterwordspacing
Y.~Gao, S.~Jia, D.~Li, C.~Huang, Z.~Meng, Y.~Wang, M.~Yu, T.~Xu, M.~Liu,
  J.~Sun, Q.~Jia, Q.~Zhang, Y.~Gao, K.~Song, X.~Wang, and Y.~Fan, ``Prediction
  model of random forest for the risk of hyperuricemia in a chinese basic
  health checkup test,'' \emph{Bioscience Reports}, vol.~41, no.~4, p.
  BSR20203859, 04 2021. [Online]. Available:
  \url{https://doi.org/10.1042/BSR20203859}
\BIBentrySTDinterwordspacing

\bibitem{llmmedicinereview}
\BIBentryALTinterwordspacing
H.~Zhou, F.~Liu, B.~Gu, X.~Zou, J.~Huang, J.~Wu, Y.~Li, S.~S. Chen, P.~Zhou,
  J.~Liu, Y.~Hua, C.~Mao, C.~You, X.~Wu, Y.~Zheng, L.~Clifton, Z.~Li, J.~Luo,
  and D.~A. Clifton, ``A survey of large language models in medicine: Progress,
  application, and challenge,'' 2024. [Online]. Available:
  \url{https://arxiv.org/abs/2311.05112}
\BIBentrySTDinterwordspacing

\bibitem{babenkoTabular}
\BIBentryALTinterwordspacing
Y.~Gorishniy, I.~Rubachev, V.~Khrulkov, and A.~Babenko, ``Revisiting deep
  learning models for tabular data,'' 2023. [Online]. Available:
  \url{https://arxiv.org/abs/2106.11959}
\BIBentrySTDinterwordspacing

\bibitem{Sampa2020}
\BIBentryALTinterwordspacing
M.~B. Sampa, M.~N. Hossain, M.~R. Hoque, R.~Islam, F.~Yokota, M.~Nishikitani,
  and A.~Ahmed, ``Blood uric acid prediction with machine learning: Model
  development and performance comparison,'' \emph{JMIR Med Inform}, vol.~8,
  no.~10, p. e18331, Oct 2020. [Online]. Available:
  \url{https://medinform.jmir.org/2020/10/e18331}
\BIBentrySTDinterwordspacing

\bibitem{Vohra2022}
\BIBentryALTinterwordspacing
R.~Vohra, A.~Hussain, A.~K. Dudyala, J.~Pahareeya, and W.~Khan, ``Multi-class
  classification algorithms for the diagnosis of anemia in an outpatient
  clinical setting,'' \emph{PLOS ONE}, vol.~17, no.~7, pp. 1--18, 07 2022.
  [Online]. Available: \url{https://doi.org/10.1371/journal.pone.0269685}
\BIBentrySTDinterwordspacing

\bibitem{emmanuel2021survey}
T.~Emmanuel, T.~Maupong, D.~Mpoeleng, T.~Semong, B.~Mphago, and O.~Tabona, ``A
  survey on missing data in machine learning,'' \emph{Journal of Big data},
  vol.~8, pp. 1--37, 2021.

\bibitem{joel2022review}
L.~O. Joel, W.~Doorsamy, and B.~S. Paul, ``A review of missing data handling
  techniques for machine learning,'' \emph{International Journal of Innovative
  Technology and Interdisciplinary Sciences}, vol.~5, no.~3, pp. 971--1005,
  2022.

\bibitem{katoch2021review}
S.~Katoch, S.~S. Chauhan, and V.~Kumar, ``A review on genetic algorithm: past,
  present, and future,'' \emph{Multimedia tools and applications}, vol.~80, pp.
  8091--8126, 2021.

\bibitem{sulejmani2021development}
A.~Sulejmani and O.~Ko{\c{c}}a, ``Development of optimal transmission rate of
  the kinematic chain by using genetic algorithms coded in mathcad,''
  \emph{International Journal of Innovative Technology and Interdisciplinary
  Sciences}, vol.~4, no.~4, pp. 792--803, 2021.

\bibitem{yoon2020gamin}
S.~Yoon and S.~Sull, ``Gamin: Generative adversarial multiple imputation
  network for highly missing data,'' in \emph{Proceedings of the IEEE/CVF
  conference on computer vision and pattern recognition}, 2020, pp. 8456--8464.

\bibitem{kim2020survey}
J.~Kim, D.~Tae, and J.~Seok, ``A survey of missing data imputation using
  generative adversarial networks,'' in \emph{2020 International conference on
  artificial intelligence in information and communication (ICAIIC)}.\hskip 1em
  plus 0.5em minus 0.4em\relax IEEE, 2020, pp. 454--456.

\bibitem{el2019machine}
E.~El-Kenawy and E.~SM, ``A machine learning model for hemoglobin estimation
  and anemia classification,'' \emph{International Journal of Computer Science
  and Information Security (IJCSIS)}, vol.~17, no.~2, pp. 100--108, 2019.

\bibitem{SetTransformer}
\BIBentryALTinterwordspacing
J.~Lee, Y.~Lee, J.~Kim, A.~R. Kosiorek, S.~Choi, and Y.~W. Teh, ``Set
  transformer: A framework for attention-based permutation-invariant neural
  networks,'' 2019. [Online]. Available: \url{https://arxiv.org/abs/1810.00825}
\BIBentrySTDinterwordspacing

\bibitem{Vaswany}
A.~{V}aswani, N.~{S}hazeer, N.~{P}armar, J.~{U}szkoreit, L.~{J}ones, A.~N.
  {G}omez, {\L}.~{K}aiser, and I.~{P}olosukhin, ``{A}ttention is {A}ll you
  {N}eed,'' in \emph{Advances in Neural Information Processing Systems},
  I.~Guyon, U.~V. Luxburg, S.~Bengio, H.~Wallach, R.~Fergus, S.~Vishwanathan,
  and R.~Garnett, Eds., vol.~30.\hskip 1em plus 0.5em minus 0.4em\relax Curran
  Associates, Inc., 2017.

\bibitem{yandexgpt}
\BIBentryALTinterwordspacing
 [Online]. Available: \url{https://yandex.cloud/en-ru/services/yandexgpt}
\BIBentrySTDinterwordspacing

\bibitem{TabTransformer}
\BIBentryALTinterwordspacing
X.~Huang, A.~Khetan, M.~Cvitkovic, and Z.~Karnin, ``Tabtransformer: Tabular
  data modeling using contextual embeddings,'' 2020. [Online]. Available:
  \url{https://arxiv.org/abs/2012.06678}
\BIBentrySTDinterwordspacing

\bibitem{keras}
F.~Chollet \emph{et~al.}, ``Keras,'' \url{https://keras.io}, 2015.

\bibitem{tensorflow}
\BIBentryALTinterwordspacing
M.~Abadi, A.~Agarwal, P.~Barham, E.~Brevdo, Z.~Chen, C.~Citro, G.~S. Corrado,
  A.~Davis, J.~Dean, M.~Devin, S.~Ghemawat, I.~Goodfellow, A.~Harp, G.~Irving,
  M.~Isard, Y.~Jia, R.~Jozefowicz, L.~Kaiser, M.~Kudlur, J.~Levenberg,
  D.~Man\'{e}, R.~Monga, S.~Moore, D.~Murray, C.~Olah, M.~Schuster, J.~Shlens,
  B.~Steiner, I.~Sutskever, K.~Talwar, P.~Tucker, V.~Vanhoucke, V.~Vasudevan,
  F.~Vi\'{e}gas, O.~Vinyals, P.~Warden, M.~Wattenberg, M.~Wicke, Y.~Yu, and
  X.~Zheng, ``{TensorFlow}: Large-scale machine learning on heterogeneous
  systems,'' 2015, software available from tensorflow.org. [Online]. Available:
  \url{https://www.tensorflow.org/}
\BIBentrySTDinterwordspacing

\bibitem{MultiTask}
\BIBentryALTinterwordspacing
S.~Sosnin, M.~Vashurina, M.~Withnall, P.~Karpov, M.~Fedorov, and I.~V. Tetko,
  ``A survey of multi-task learning methods in chemoinformatics,''
  \emph{Molecular Informatics}, vol.~38, no.~4, p. 1800108, 2019. [Online].
  Available:
  \url{https://onlinelibrary.wiley.com/doi/abs/10.1002/minf.201800108}
\BIBentrySTDinterwordspacing

\bibitem{BabenkoEmbeddings}
\BIBentryALTinterwordspacing
Y.~Gorishniy, I.~Rubachev, and A.~Babenko, ``On embeddings for numerical
  features in tabular deep learning,'' 2023. [Online]. Available:
  \url{https://arxiv.org/abs/2203.05556}
\BIBentrySTDinterwordspacing

\bibitem{ensemble}
\BIBentryALTinterwordspacing
Y.~Gorishniy, A.~Kotelnikov, and A.~Babenko, ``Tabm: Advancing tabular deep
  learning with parameter-efficient ensembling,'' 2025. [Online]. Available:
  \url{https://arxiv.org/abs/2410.24210}
\BIBentrySTDinterwordspacing

\bibitem{federated}
\BIBentryALTinterwordspacing
B.~Yurdem, M.~Kuzlu, M.~K. Gullu, F.~O. Catak, and M.~Tabassum, ``Federated
  learning: Overview, strategies, applications, tools and future directions,''
  \emph{Heliyon}, vol.~10, no.~19, p. e38137, 2024. [Online]. Available:
  \url{https://www.sciencedirect.com/science/article/pii/S2405844024141680}
\BIBentrySTDinterwordspacing

\end{thebibliography}

\end{document}